\documentclass[nohyperref]{article}

\usepackage{microtype}
\usepackage{graphicx}
\usepackage{subfigure}
\usepackage{booktabs} 

\usepackage{hyperref}

\usepackage{adjustbox}
\usepackage{float}

\usepackage{multirow}
\usepackage[table,xcdraw]{xcolor}

\usepackage[accepted]{icml2022}

\usepackage{amssymb}
\usepackage{mathtools}
\usepackage{amsthm}

\newcommand\crule[3][black]{\textcolor{#1}{\rule{#2}{#3}}}
\definecolor{myorange}{rgb}{1, 0.396, 0.235}
\definecolor{myblue}{rgb}{0, 0.424, 0.706}
\definecolor{myred}{rgb}{0.773, 0.176, 0.071}

\definecolor{fig2_g}{rgb}{0.23, 0.64, 0.31}
\definecolor{fig2_r}{rgb}{0.85, 0.22, 0.16}

\usepackage[capitalize,noabbrev]{cleveref}

\theoremstyle{plain}
\newtheorem{theorem}{Theorem}[section]
\newtheorem{theorem*}{Theorem}[section]

\theoremstyle{definition}

\theoremstyle{remark}

\usepackage[textsize=tiny]{todonotes}

\icmltitlerunning{On Regularization Effects of Quantization}

\begin{document}

\twocolumn[
\icmltitle{QReg: On Regularization Effects of Quantization}



\icmlsetsymbol{equal}{*}

\begin{icmlauthorlist}
\icmlauthor{MohammadHossein AskariHemmat}{polymtl}
\icmlauthor{Reyhane Askari Hemmat}{mila}
\icmlauthor{Alex Hoffman}{Deeplite}
\icmlauthor{Ivan Lazarevich}{Deeplite}
\icmlauthor{Ehsan Saboori}{Deeplite}
\icmlauthor{Olivier Mastropietro}{Deeplite}
\icmlauthor{Sudhakar Sah}{Deeplite}
\icmlauthor{Yvon Savaria}{polymtl}
\icmlauthor{Jean-Pierre David}{polymtl}
\end{icmlauthorlist}

\icmlaffiliation{polymtl}{Ecole Polytechnique Montreal}
\icmlaffiliation{mila}{Mila, Université de Montreal }
\icmlaffiliation{Deeplite}{Deeplite Inc., Montreal}

\icmlcorrespondingauthor{MohammadHossein AskariHemmat}{m.h.askari.hemmat@gmail.com}

\icmlkeywords{Machine Learning, ICML}

\vskip 0.3in
]



\printAffiliationsAndNotice{}  

\begin{abstract}
In this paper we study the effects of quantization in DNN training. We hypothesize that weight quantization is a form of regularization and the amount of regularization is correlated with the quantization level (precision). We confirm our hypothesis by providing analytical study and empirical results. By modeling weight quantization as a form of additive noise to weights, we explore how this noise propagates through the network at training time. We then show that the magnitude of this noise is correlated with the level of quantization. To confirm our analytical study, we performed an extensive list of experiments summarized in this paper in which we show that the regularization effects of quantization can be seen in various vision tasks and models, over various datasets. Based on our study, we propose that 8-bit quantization provides a reliable form of regularization in different vision tasks and models.  
\end{abstract}

\section{Introduction}




Deep neural network (DNN) quantization has been an active research topic in the past few years. Quantization is the  task of approximating  high precision (floating point) weights and activations of a model with lower bit resolution counterparts (usually 1 to 8 bit integers). Quantization can be done after (post training quantization) or during training (quantization aware training). Although both methods can compress the model and reduce overall execution time, at the cost of model accuracy, post training quantization usually takes less time to quantize a model. On the other hand, quantization aware training usually requires a full training session and with a moderate quantization level (int8) can produce quantized models with little to no accuracy loss compared to their full precision counterparts. In fact, as we will discuss in Section \ref{sec:qreg_background}, there are many existing works claiming that quantization aware training produced accuracy improvements compared to some of the best full precision models. In these studies, the authors usually attribute such accuracy gains to the regularization effect of quantization. This has motivated us to explore how quantization affects training. In this work, we first hypothesize the regularization effect of quantization. Moreover, we hypothesize that this regularization effect is correlated with the quantization level. To confirm our hypothesis, we analytically explore how quantization noise propagates in a network at training time, and how the quantization level is correlated with the amount of regularization. We then empirically explore the regularization effect of quantization by applying different quantization aware training methods on different models and datasets with different quantization levels. In both analytical and empirical studies, we confirm our initial hypothesis.

\begin{figure}[!t]
\label{fig:yolo_overfit}
\centering
  \begin{picture}(300,140)
    \put(-15,20){\rotatebox{90}{Objectiveness Loss Value}}
    \put(100,-10){Epoch Number}
    \includegraphics[width=8cm]{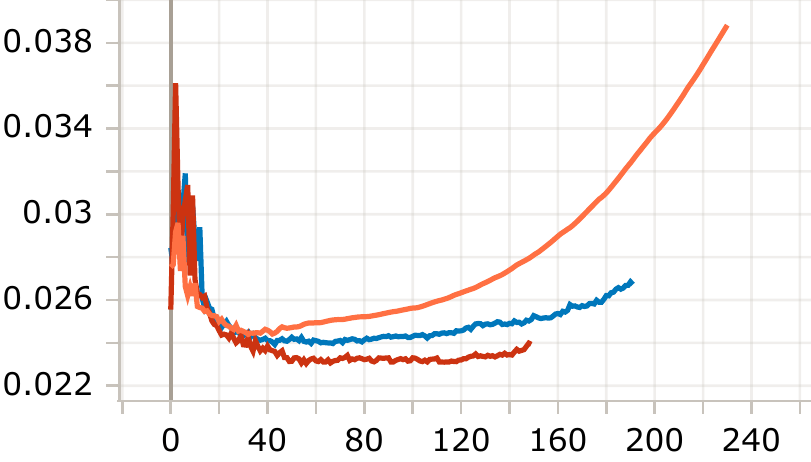}
    \end{picture}
\caption{Validation objective loss of 4-bit quantized \crule[myred]{0.2cm}{0.2cm}, 8-bit quantized \crule[myblue]{0.2cm}{0.2cm} and full precision model \crule[myorange]{0.2cm}{0.2cm} for the YOLOv5n model on the VOC dataset. This figure illustrates that unlike the quantized model, the full precision model exhibits heavy overfitting. We believe that the quantized model has better generalization performance due to its regularization effect.}
\vspace{-1.6cm}
\end{figure}

\section{Background}
\label{sec:qreg_background}

The regularization effect of quantization has previously been reported in the literature. We have categorized earlier works into three main lines of work, which we will review briefly in the following subsections.
\subsection{Effect of Quantization on Accuracy Improvements}
\label{sec:effect_of_quant_acc_improv}
In \cite{NIPS2015_3e15cc11}, one of the first quantization methods for DNNs, the authors empirically show that modern deep learning optimization techniques such as stochastic gradient descent are compatible with noisy weights, as they appear in quantization. They also empirically show that noisy weights provide a form of regularization. They confirm their hypothesis by achieving state-of-the-art results on MNIST, CIFAR-10 and SVHN datasets. Similar to this work, authors in \cite{10.5555/3122009.3242044}, argue that weight quantization is a form of noise injection to weights and as shown in DropConnect \cite{hou2018loss}, this noise injection can improve the generalization performance.
However, both studies are only limited to small-scale classification tasks. Their studies are also limited to one quantization technique. Furthermore, their hypotheses lack analytical study.


In \cite{abdolrashidi2021pareto} the authors study the effect of quantization on accuracy and provide an analysis of the regularization effect of quantization. Their study is limited to showing a generalization gap between a quantized model with 8 and 4 bit precision and full precision models on the Imagenet dataset on Resnet50. On the other hand, in addition to their work, they reported the generalization gap  of different quantization methods. However, in their experiments, the regularization effect of quantization is not investigated.
Furthermore, they did not explain whether this generalization gap is coming from quantization or whether they have used other regularization techniques that might have affected the overall results.

Apart from the papers above, there are many other works reporting accuracy improvement after quantization, \cite{DBLP:conf/iclr/XuYLOCWZ18,louizos2018relaxed,10.1145/3451211, xu2018quantization,7878541,DBLP:journals/corr/abs-1904-04402,9047891,8999090}. However, to the best of our knowledge, none of them provided analytical or empirical studies and the authors justified the boost in performance (accuracy) with a conjectured regularization effect of quantization.

\subsection{Analytical Studies}
\label{sec:qreg_analytical_study}
As presented in \ref{sec:effect_of_quant_acc_improv}, the regularization effect of quantization has been observed in previous works. In the following, we present three papers that provide an analytical study on how quantization affects training.

In \cite{https://doi.org/10.48550/arxiv.1606.01981}, the authors claim that applying a constraint on the weights can act as a regularizer. They investigate how DNNs are robust to different types of weight distortion. As an example, for quantization, they studied the effect of binarization proposed in \cite{NIPS2015_3e15cc11} on training. Using proximal methods, the authors showed that applying weight clipping during  training can be modeled as a type of regularization that penalizes weights outside the unit ball of the $\ell_\propto$ norm. Their study is limited to applying binarization on weights and they do not explain how different levels of weight quantization affect training. Moreover, like \cite{NIPS2015_3e15cc11} , they only investigate the regularization effect of weight distortion on small models.

A more recent study in  \cite{Alizadeh2020Gradient} provides an analytical overview on how quantization can affect robustness. Although not directly relevant to our work, the authors in this paper provide an analytical study on how quantization affects training and the loss function. In this paper, the authors propose a regularization based quantization method that, unlike most quantization-aware methods, provides a selectable quantization bit-width after training. Like previous methods, the authors model quantization as a form of noise. Using first order Taylor approximation, they then provide an analytical study on how this noise affects the robustness of DNNs. With their method, they show that if the $\ell_1$-norm of the gradient is small, the perturbation (at least to the first-order approximation) can be effectively controlled for various quantization bit-widths. To guarantee the $\ell_1$-norm of the gradients, they propose a regularization term in the loss function. As we will show in \ref{sec:qreg_analytical}, we used similar quantization noise modeling and approximation techniques to investigate the effect of quantization on the loss function. 

In \cite{https://doi.org/10.48550/arxiv.2206.05916}, the authors provide an analytical study on how models with stochastic binary quantization can have a smaller generalization gap compared to their full precision counterparts. In this work, the authors propose a so-called "quasi neural network" to approximate the effect of binarization on neural networks. The authors then derive the neural tangent kernel for the proposed quasi neural network approximation. With this formalization, the authors show that binary neural networks have lower capacity, hence lower training accuracy, but also a smaller generalization gap than their full precision counterparts. Finally, the authors provide some empirical results to support their derivations. However, the authors fail to show the effect of different quantization levels on the generalization gap. As mentioned earlier, the authors only focused their analysis on statistical binary quantization and they did not explore how different quantization methods and precision levels affect generalization. Also, their empirical results are only limited to a two layer model.

\subsection{Using Quantization for its Regularization Effect}
\label{sec:qreg_analytical_study}
Although very limited, the idea of using quantization solely for its regularization effect has been explored before. In \cite{hou2017loss}, the authors studied the effect of binarization on the loss function. They then proposed an algorithm that uses diagonal Hessian approximation that directly minimizes the loss function with regard to the binarized weights. For training their binarized model, the authors claimed that they only used regularization that is induced by weight binarization. Furthermore they further developed their method in \cite{hou2018loss} where they show that the same techniques can be applied to weight ternerization. 

\section{Quantization Noise as a Regularizer}
\label{sec:qreg_analytical}

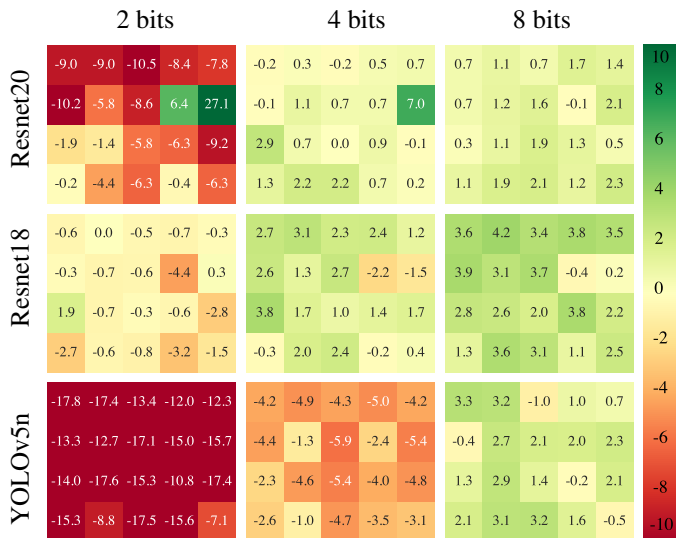
\begin{figure}[t]





  \begin{picture}(0,200)
    \put(30,198){2 bits}
    \put(110,198){4 bits}
    \put(180,198){8 bits}
    \put(-10,145){\rotatebox{90}{Resnet20}}
    \put(-10,85){\rotatebox{90}{Resnet18}}
    \put(-10,10){\rotatebox{90}{YOLOv5n}}
    \includegraphics[width=\textwidth/2]{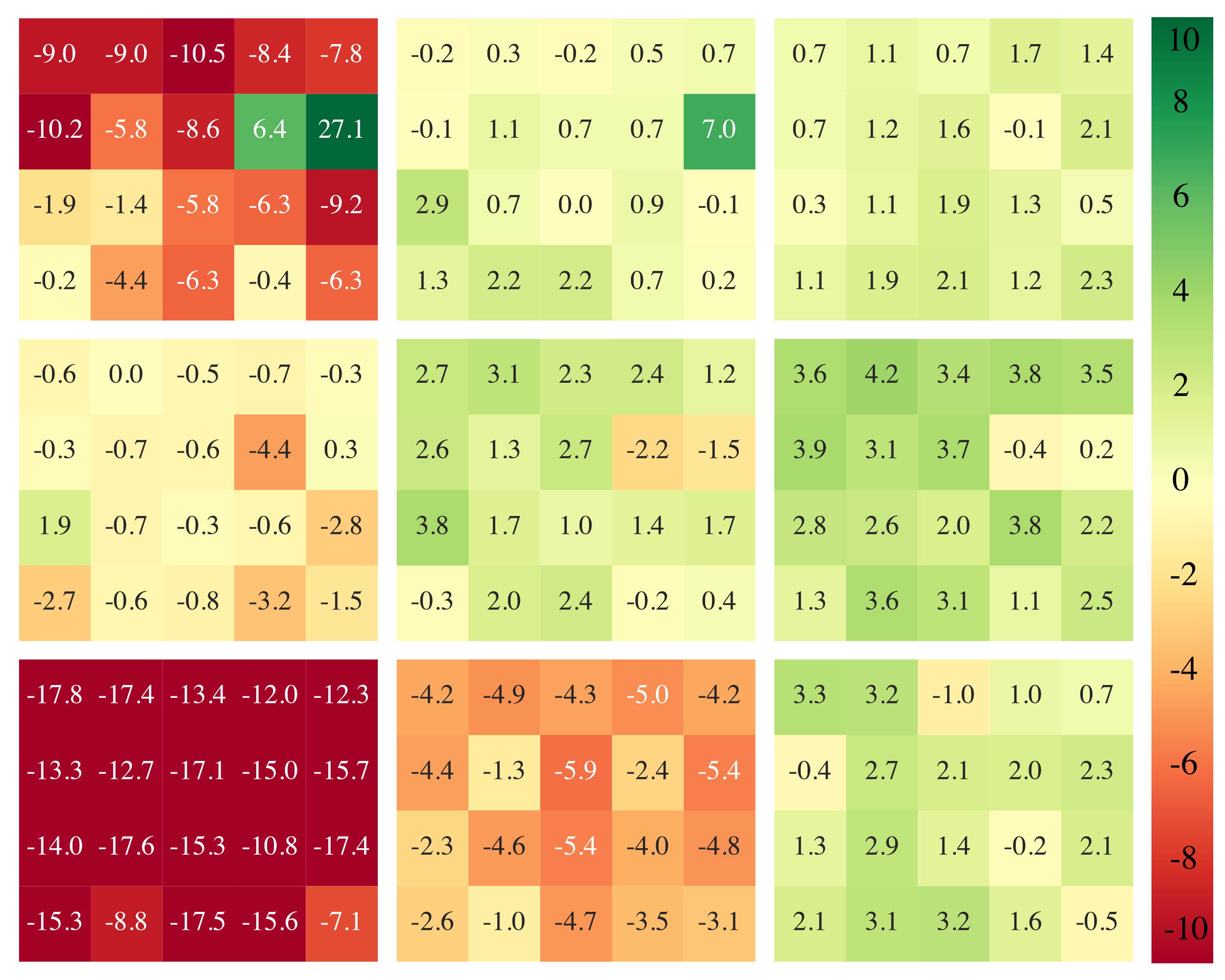}
    \end{picture}
\vspace{-0.5cm}
  \caption{Regularization effect of quantization over different quantization level (2 bits, 4 bits and 8 bits), and models. The value in each cell shows the test accuracy difference between the quantized and full precision models. The position of each cell shows the corresponding augmentation in Figure \ref{fig:aug_map} that was applied on test data. Cells are colored by generalization difference, where green cells \crule[fig2_g]{0.2cm}{0.2cm} show quantized model generalized better compared to floating point model and the red cells \crule[fig2_r]{0.2cm}{0.2cm} show the opposite. The color intensity shows the amplitude of the difference according to the color bar shown on the right.
  }
  
\label{fig:result}
\end{figure}

The authors in the previous works usually assume that weight quantization is similar to adding noise to a model's weights. However, it is not clear how this noise propagates through the network  at training time. Also, it is not clear how different quantization levels and algorithms affect the magnitude of this noise. 
In this section, we will show how quantization noise can be modeled as a regularization term. We will then extend our study and show that this noise is directly related to the quantization level. 

For simplicity, let us consider a regression problem where the mean square error loss is defined as,
\vspace{-0.5cm}

\begin{equation}\label{eq:mse}
    \mathcal{L} = \frac{1}{m} \sum_{i=1}^m \|\hat{y_i} - y_i\|_2^2,
\end{equation}
where $y_i$ is the target, and $\hat{y_i} = f(x, w)$ is the predicted target of the network $f$ parameterized by $w$. Weight quantization can be characterized as noise perturbing the weights and activations. Such noise can be described as:
\begin{align} \label{eq:quantization_noise}
    f(x,w_q) = f(x, w+\delta)
\end{align}

Thus, a quantized neural network, effectively has the following loss,
\begin{equation}\label{eq:mse_quant}
\Tilde{\mathcal{L}} = \frac{1}{m} \sum_{i=1}^m \|\hat{y_i}^q - y_i\|_2^2,    
\end{equation}
where $\hat{y_i}^q = f(x, w+\delta)$. 


\begin{table}[t]
\begin{adjustbox}{max width=8.2cm}
\begin{tabular}{c|c|c|c|}
\cline{2-4}
\multicolumn{1}{l|}{} & \begin{tabular}[c]{@{}c@{}}Quantization\\  Level\end{tabular} & \begin{tabular}[c]{@{}c@{}}Avg Accuracy\\Augmented Data\end{tabular} & \begin{tabular}[c]{@{}c@{}} Relative Error \\Improvement Eq.\ref{eq:gen_metric} \end{tabular} \\ \hline
\multicolumn{1}{|c|}{} & 2 bits & 79.17 & \cellcolor[HTML]{D83728}-8.32 \\ \cline{2-4} 
\multicolumn{1}{|c|}{} & 4 bits & 83.94 & \cellcolor[HTML]{3BA250}2.98 \\ \cline{2-4} 
\multicolumn{1}{|c|}{} & 8 bits & 84.07 & \cellcolor[HTML]{3BA250}3.33 \\ \cline{2-4} 
\multicolumn{1}{|c|}{\multirow{-4}{*}{\begin{tabular}[c]{@{}c@{}}Resnet20\\  Cifar10\end{tabular}}} & FP32 & 82.80 & 0.00 \\ \hline
\multicolumn{1}{|c|}{} & 2 bits & 49.43 & \cellcolor[HTML]{D83728}-0.84 \\ \cline{2-4} 
\multicolumn{1}{|c|}{} & 4 bits& 51.76 & \cellcolor[HTML]{3BA250}1.21 \\ \cline{2-4} 
\multicolumn{1}{|c|}{} & 8 bits & 53.05 & \cellcolor[HTML]{3BA250}2.39 \\ \cline{2-4} 
\multicolumn{1}{|c|}{\multirow{-4}{*}{\begin{tabular}[c]{@{}c@{}}Resnet18\\ Cifar100\end{tabular}}} & FP32 & 50.40 & 0.00 \\ \hline
\multicolumn{1}{|c|}{} & 2 bits& 14.66 & \cellcolor[HTML]{D83728}-7.86 \\ \cline{2-4} 
\multicolumn{1}{|c|}{} & 4 bits& 24.90 & \cellcolor[HTML]{D83728}-2.31 \\ \cline{2-4} 
\multicolumn{1}{|c|}{} & 8 bits & 30.34 & \cellcolor[HTML]{3BA250}0.96 \\ \cline{2-4} 
\multicolumn{1}{|c|}{\multirow{-4}{*}{\begin{tabular}[c]{@{}c@{}}YOLOv5n\\ VOC\end{tabular}}} & FP32 & 28.78 & 0.00 \\ \hline
\end{tabular}
\end{adjustbox}

  \caption{ Regularization effect of quantization measured with relative error improvement. For each model, we measured the test accuracy of quantized (2-bit, 4-bit and 8-bit) and full precision (FP32) models when 19 different data augmentation methods (presented in \ref{fig:aug_map}) are applied on the original test dataset. We then averaged the results (third column). For each case (row), we used Equation \ref{eq:gen_metric} to measure the  relative test error improvement. Green cells \crule[fig2_g]{0.2cm}{0.2cm} in the last column show that quantized models have less error compared to the full precision model. The red cells \crule[fig2_r]{0.2cm}{0.2cm} show the opposite.}
  
\label{tab:t_result}
\end{table}

\begin{theorem}\label{th:regression}
Given the mean square error loss defined in Eq. \ref{eq:mse_quant}, assuming that the quantization noise follows a normal distribution, $\delta \sim \mathcal{N} (0, \sigma I)$, applying a first-order Taylor approximation around the weights of the full precision model, results in the following approximation on the $\Tilde{\mathcal{L}}$:
\vspace{-0.5cm}

\begin{equation}\label{eq:mse_2}
    \Tilde{\mathcal{L}} \approx \mathcal{L} + \frac{\sigma \delta^2}{m} \sum_{i=1}^m  \|\nabla_w \hat{y_i} \|^2_2,
\end{equation}
This means that minimizing $\Tilde{\mathcal{L}}$ is equivalent to minimizing the loss of a non-quantized neural network with gradient norm regularization.
\end{theorem}

Derivation in Appendix \ref{appendix:a}.

Direct application of noise on the weights as a form of regularization has been studied in the literature. In \cite{goodfellow2016deep} authors argue that noisy weights result in forcing the model to find minima that are insensitive to small perturbations of the weights. In other words, noisy weights encourage finding minima in flat regions and thus lead to better generalization \cite{hochreiter1994simplifying}.





\section{Experimental Studies}
In this section, we will describe empirically how quantization has a regularization effect. Unlike previous works that was described in Section \ref{sec:qreg_background}, we tried to explore the regularization effect of quantization over different quantization techniques, levels, models, vision tasks and datasets. In the following subsections, we briefly describe our test setup and results.

\subsection{Experiment Setup}
\subsubsection{Training Setup}
\label{sec:train_setup}
As we described in Section \ref{sec:qreg_analytical} we hypothesize that quantization has a regularization effect that is correlated with quantization level (precision). In order to test this hypothesis, we needed a quantization technique that could be used to quantize a model to any quantization precision. For our experiments, we picked Learned Step size  Quantization (LSQ) \cite{Esser2020LEARNED} and Parameterized Clipping Activation (PACT) \cite{choi2018pact}. Since PACT applies quantization only to activations, we used DoReFa \cite{zhou2016dorefa} quantization for weights. For PACT, DoReFa and LSQ, we used learning rate of 0.1 with momentum of 0.9 for the classification experiment. In our experiments, we initially used quantization as the only regularizer in the network. However, both full precision and quantized models were unable to achieve a good accuracy. Hence, we used weight decay of 0.0001 for both full precision and quantized models in our experiments. We trained both full precision and quantized models for 200 epochs. We used a multi-step learning scheduler with milestones of 50, 90, 130 and 180 with gamma of 0.2.

\subsubsection{Models and Datasets}
\label{sec:model_gen_gap}
In our experiments, we explored the regularization effect of quantization in different vision tasks, models, quantization levels and datasets. We explored two popular vision tasks, object detection and classification. For small-scale classification tasks, we tested our hypothesis on CIFAR10 dataset and used Resnet20, Mobilenet V1, Resnet18 and Resnet50. For bigger classification tasks we used CIFAR100 dataset and used Mobilenet V1, Resnet18, Resnet50. For object detection, we used VOC dataset and tested our hypothesis on YOLOv5n model. The full list of experiments is available in Appendix \ref{appendix:b}. 
For each experiment, we trained the model as described in \ref{sec:train_setup}. To measure the generalization improvement, in addition to the original test data, we applied 19 different augmentations on the original dataset as provided in \cite{hendrycks2019robustness}. We then tested each full precision and quantized model with the aforementioned datasets. To measure generalization improvement, we calculated the test accuracy difference between the full precision model and the quantized model when an augmentation is applied on the corresponding test dataset.

\subsubsection{Results}
Figure \ref{fig:result} illustrates part of the experiments that we ran to test our hypothesis. Figure \ref{fig:result} shows results of nine different test configuration in nine squares. Each square consists of 20 cells, which as we discussed in Section \ref{sec:model_gen_gap}, corresponds to the test accuracy difference between full precision model and quantized models when different augmentation is used. From left to right, each column shows the quantization levels (2-bit, 4-bit and 8-bit) that was used for our tests. From top to bottom, we have tested our hypothesis on Resnet20 on cifar10, Resnet18 on cifar100 and YOLOv5n on VOC dataset respectively. As it can be seen (i.e. the overall cell colors in the right column squares are green), regardless of model, dataset, augmentation  or quantization technique, 8-bit quantized models generally have better generalization compared to full precision models. As we discussed in \ref{sec:qreg_analytical_study}, we believe that this performance improvement corresponds to the regularization effect of quantizing the models. As we predicted, this regularization effect is correlated with the quantization level. Moving from left to right columns, the overall color codes for each cell changes from red to green. Which indicates that quantized models are generalizing better compared to their full precision counterparts as we use higher precision quantized models (from 2-bit to 8-bit).  Regardless of the model, dataset, data augmentation  or quantization technique, the generalization difference shrinks and even get worse as we use less precision for quantization. Another way to evaluate how quantization is helping with better generalization, is to compare the error of quantized and full precision models.
Table \ref{tab:t_result} shows the performance improvement relative to model error. We averaged the accuracy of each square in Figure \ref{fig:result}. We then, calculated the performance improvement relative to error using the following formula:
\begin{equation}
\label{eq:gen_metric}
    Error Improvement = log\Big(\frac{100-fval}{100-qval}\Big)*100
\end{equation}
Where $fval$ and $qval$ are the average accuracy of full precision and quantized model over 19 different augmentation setups. Since models on small datasets like cifar10 generally have small errors compared to models on more complex datasets (like cifar100 or VOC), we used a logarithm in Equation \ref{eq:gen_metric}. Once again, according to Table \ref{tab:t_result}, in all cases 8-bit quantized models reduce the error more (i.e, generalizing better) compared to full precision models. On all three models tested over 19 different augmentation setups, all 8-bit quantized models reduce the error when compared to full precision models (i.e last column for 8-bit quantization levels is green for all models). This effect (reducing error) is less clear as we use lower precision models .

Our results presented in Figure \ref{fig:result}, Table \ref{tab:t_result} and Appendix \ref{appendix:b} empirically confirm our hypothesis. Unlike \cite{NIPS2015_3e15cc11}, we believe that the regularization effect of quantization is in fact correlated with the quantization level. Our empirical study shows that moderate quantization (8-bit) generally helps models generalize better. In addition to these results, specially for more complex models, we observe that quantized models overfit less at training time. Figure \ref{fig:yolo_overfit} illustrates the validation loss of 8-bit, 4-bit quantized and full precision for the YOLOv5n model applied to the VOC dataset. Due to regularization effect of quantization, the quantized models overfit much less compared to full precision model. This, confirms our hypothesis. For the full list of experiments, please refer to Appendix \ref{appendix:b}.


\section{Conclusion}

In this paper, we explored the regularization effect of quantization. We hypothesized that this regularization effect is correlated with the quantization level. To confirm our hypothesis, we provided an analytical study as well as an extensive list of experiments. Our experiments show that the regularization effect of quantization exist regardless of model, dataset, vision task and quantization technique. Finally, we analytically and empirically confirm that this regularization effect is correlated with quantization level. Our empirical study shows that moderate quantization (8-bit) helps models generalize better.

\section*{Acknowledgements}
The authors would like to thank Mohammad Pezeshki and Anush Sankaran for helpful discussion about designing empirical studies presented in this paper. The authors would also like to acknowledge the support for this project from Fonds de Recherche du Québec Nature et technologies (FRQNT) and Natural Sciences and Engineering Research Council of Canada (NSERC).

\bibliography{references}

\begin{thebibliography}{22}
\providecommand{\natexlab}[1]{#1}
\providecommand{\url}[1]{\texttt{#1}}
\expandafter\ifx\csname urlstyle\endcsname\relax
  \providecommand{\doi}[1]{doi: #1}\else
  \providecommand{\doi}{doi: \begingroup \urlstyle{rm}\Url}\fi

\bibitem[Abdolrashidi et~al.(2021)Abdolrashidi, Wang, Agrawal, Malmaud,
  Rybakov, Leichner, and Lew]{abdolrashidi2021pareto}
Abdolrashidi, A., Wang, L., Agrawal, S., Malmaud, J., Rybakov, O., Leichner,
  C., and Lew, L.
\newblock Pareto-optimal quantized resnet is mostly 4-bit.
\newblock In \emph{Proceedings of the IEEE/CVF Conference on Computer Vision
  and Pattern Recognition}, pp.\  3091--3099, 2021.

\bibitem[Alizadeh et~al.(2020)Alizadeh, Behboodi, van Baalen, Louizos,
  Blankevoort, and Welling]{Alizadeh2020Gradient}
Alizadeh, M., Behboodi, A., van Baalen, M., Louizos, C., Blankevoort, T., and
  Welling, M.
\newblock Gradient $\ell_1$ regularization for quantization robustness.
\newblock In \emph{International Conference on Learning Representations}, 2020.
\newblock URL \url{https://openreview.net/forum?id=ryxK0JBtPr}.

\bibitem[Andri et~al.(2018)Andri, Cavigelli, Rossi, and Benini]{7878541}
Andri, R., Cavigelli, L., Rossi, D., and Benini, L.
\newblock Yodann: An architecture for ultralow power binary-weight cnn
  acceleration.
\newblock \emph{IEEE Transactions on Computer-Aided Design of Integrated
  Circuits and Systems}, 37\penalty0 (1):\penalty0 48--60, 2018.
\newblock \doi{10.1109/TCAD.2017.2682138}.

\bibitem[Chen et~al.(2021)Chen, Qiu, Zhuang, Zhang, Hu, Lu, Wang, Shi, Huang,
  and Xu]{10.1145/3451211}
Chen, W., Qiu, H., Zhuang, J., Zhang, C., Hu, Y., Lu, Q., Wang, T., Shi, Y.,
  Huang, M., and Xu, X.
\newblock Quantization of deep neural networks for accurate edge computing.
\newblock \emph{J. Emerg. Technol. Comput. Syst.}, 17\penalty0 (4), jun 2021.
\newblock ISSN 1550-4832.
\newblock \doi{10.1145/3451211}.
\newblock URL \url{https://doi.org/10.1145/3451211}.

\bibitem[Choi et~al.(2018)Choi, Wang, Venkataramani, Chuang, Srinivasan, and
  Gopalakrishnan]{choi2018pact}
Choi, J., Wang, Z., Venkataramani, S., Chuang, P. I.-J., Srinivasan, V., and
  Gopalakrishnan, K.
\newblock {PACT}: Parameterized clipping activation for quantized neural
  networks, 2018.
\newblock URL \url{https://openreview.net/forum?id=By5ugjyCb}.

\bibitem[Courbariaux et~al.(2015)Courbariaux, Bengio, and
  David]{NIPS2015_3e15cc11}
Courbariaux, M., Bengio, Y., and David, J.-P.
\newblock Binaryconnect: Training deep neural networks with binary weights
  during propagations.
\newblock In Cortes, C., Lawrence, N., Lee, D., Sugiyama, M., and Garnett, R.
  (eds.), \emph{Advances in Neural Information Processing Systems}, volume~28.
  Curran Associates, Inc., 2015.
\newblock URL
  \url{https://proceedings.neurips.cc/paper/2015/file/3e15cc11f979ed25912dff5b0669f2cd-Paper.pdf}.

\bibitem[Esser et~al.(2020)Esser, McKinstry, Bablani, Appuswamy, and
  Modha]{Esser2020LEARNED}
Esser, S.~K., McKinstry, J.~L., Bablani, D., Appuswamy, R., and Modha, D.~S.
\newblock Learned step size quantization.
\newblock In \emph{International Conference on Learning Representations}, 2020.
\newblock URL \url{https://openreview.net/forum?id=rkgO66VKDS}.

\bibitem[Goodfellow et~al.(2016)Goodfellow, Bengio, and
  Courville]{goodfellow2016deep}
Goodfellow, I., Bengio, Y., and Courville, A.
\newblock \emph{Deep learning}.
\newblock MIT press, 2016.

\bibitem[Hendrycks \& Dietterich(2019)Hendrycks and
  Dietterich]{hendrycks2019robustness}
Hendrycks, D. and Dietterich, T.
\newblock Benchmarking neural network robustness to common corruptions and
  perturbations.
\newblock \emph{Proceedings of the International Conference on Learning
  Representations}, 2019.

\bibitem[Hochreiter \& Schmidhuber(1994)Hochreiter and
  Schmidhuber]{hochreiter1994simplifying}
Hochreiter, S. and Schmidhuber, J.
\newblock Simplifying neural nets by discovering flat minima.
\newblock \emph{Advances in neural information processing systems}, 7, 1994.

\bibitem[Hou \& Kwok(2018)Hou and Kwok]{hou2018loss}
Hou, L. and Kwok, J.~T.
\newblock Loss-aware weight quantization of deep networks.
\newblock In \emph{International Conference on Learning Representations}, 2018.

\bibitem[Hou et~al.(2017)Hou, Yao, and Kwok]{hou2017loss}
Hou, L., Yao, Q., and Kwok, J.~T.
\newblock Loss-aware binarization of deep networks.
\newblock In \emph{International Conference on Learning Representations}, 2017.

\bibitem[Hubara et~al.(2017)Hubara, Courbariaux, Soudry, El-Yaniv, and
  Bengio]{10.5555/3122009.3242044}
Hubara, I., Courbariaux, M., Soudry, D., El-Yaniv, R., and Bengio, Y.
\newblock Quantized neural networks: Training neural networks with low
  precision weights and activations.
\newblock \emph{J. Mach. Learn. Res.}, 18\penalty0 (1):\penalty0 6869–6898,
  jan 2017.
\newblock ISSN 1532-4435.

\bibitem[Liu et~al.(2020)Liu, Zhang, Ding, Xu, Jiang, and Shi]{9047891}
Liu, J., Zhang, J., Ding, Y., Xu, X., Jiang, M., and Shi, Y.
\newblock Binarizing weights wisely for edge intelligence: Guide for partial
  binarization of deconvolution-based generators.
\newblock \emph{IEEE Transactions on Computer-Aided Design of Integrated
  Circuits and Systems}, 39\penalty0 (12):\penalty0 4748--4759, 2020.
\newblock \doi{10.1109/TCAD.2020.2983370}.

\bibitem[Louizos et~al.(2019)Louizos, Reisser, Blankevoort, Gavves, and
  Welling]{louizos2018relaxed}
Louizos, C., Reisser, M., Blankevoort, T., Gavves, E., and Welling, M.
\newblock Relaxed quantization for discretized neural networks.
\newblock In \emph{International Conference on Learning Representations}, 2019.
\newblock URL \url{https://openreview.net/forum?id=HkxjYoCqKX}.

\bibitem[Merolla et~al.(2016)Merolla, Appuswamy, Arthur, Esser, and
  Modha]{https://doi.org/10.48550/arxiv.1606.01981}
Merolla, P., Appuswamy, R., Arthur, J., Esser, S.~K., and Modha, D.
\newblock Deep neural networks are robust to weight binarization and other
  non-linear distortions, 2016.
\newblock URL \url{https://arxiv.org/abs/1606.01981}.

\bibitem[Mishchenko et~al.(2019)Mishchenko, Goren, Sun, Beauchene, Matsoukas,
  Rybakov, and Vitaladevuni]{8999090}
Mishchenko, Y., Goren, Y., Sun, M., Beauchene, C., Matsoukas, S., Rybakov, O.,
  and Vitaladevuni, S. N.~P.
\newblock Low-bit quantization and quantization-aware training for
  small-footprint keyword spotting.
\newblock In \emph{2019 18th IEEE International Conference On Machine Learning
  And Applications (ICMLA)}, pp.\  706--711, 2019.
\newblock \doi{10.1109/ICMLA.2019.00127}.

\bibitem[Wang et~al.(2019)Wang, Cai, Gao, and
  Vasconcelos]{DBLP:journals/corr/abs-1904-04402}
Wang, X., Cai, Z., Gao, D., and Vasconcelos, N.
\newblock Towards universal object detection by domain attention.
\newblock \emph{CoRR}, abs/1904.04402, 2019.
\newblock URL \url{http://arxiv.org/abs/1904.04402}.

\bibitem[Xu et~al.(2018{\natexlab{a}})Xu, Yao, Lin, Ou, Cao, Wang, and
  Zha]{DBLP:conf/iclr/XuYLOCWZ18}
Xu, C., Yao, J., Lin, Z., Ou, W., Cao, Y., Wang, Z., and Zha, H.
\newblock Alternating multi-bit quantization for recurrent neural networks.
\newblock In \emph{6th International Conference on Learning Representations,
  {ICLR} 2018, Vancouver, BC, Canada, April 30 - May 3, 2018, Conference Track
  Proceedings}. OpenReview.net, 2018{\natexlab{a}}.
\newblock URL \url{https://openreview.net/forum?id=S19dR9x0b}.

\bibitem[Xu et~al.(2018{\natexlab{b}})Xu, Lu, Yang, Hu, Chen, Hu, and
  Shi]{xu2018quantization}
Xu, X., Lu, Q., Yang, L., Hu, S., Chen, D., Hu, Y., and Shi, Y.
\newblock Quantization of fully convolutional networks for accurate biomedical
  image segmentation.
\newblock \emph{Preprint at https://arxiv. org/abs/1803.04907},
  2018{\natexlab{b}}.

\bibitem[Zhang et~al.(2022)Zhang, Yin, and
  Wang]{https://doi.org/10.48550/arxiv.2206.05916}
Zhang, K., Yin, M., and Wang, Y.-X.
\newblock Why quantization improves generalization: Ntk of binary weight neural
  networks, 2022.
\newblock URL \url{https://arxiv.org/abs/2206.05916}.

\bibitem[Zhou et~al.(2016)Zhou, Wu, Ni, Zhou, Wen, and Zou]{zhou2016dorefa}
Zhou, S., Wu, Y., Ni, Z., Zhou, X., Wen, H., and Zou, Y.
\newblock Dorefa-net: Training low bitwidth convolutional neural networks with
  low bitwidth gradients.
\newblock \emph{CoRR}, abs/1606.06160, 2016.
\newblock URL \url{http://arxiv.org/abs/1606.06160}.

\end{thebibliography}
\bibliographystyle{icml2022}

\newpage
\appendix
\onecolumn
\section{Analytical Study}
\label{appendix:a}
\subsection{Derivation of Theorem \ref{th:regression}}

\begin{theorem*}
Given the mean square error loss defined in Eq. \ref{eq:mse_quant}, assuming that $\delta \sim \mathcal{N} (0, \sigma I)$, applying a first-order Taylor approximation around the weights of the full precision model, results in the following approximation on the $\Tilde{\mathcal{L}}$:

\begin{equation}
    \Tilde{\mathcal{L}} \approx \mathcal{L} +\frac{\sigma \delta^2}{m}  \sum_{i=1}^m  \|\nabla_w \hat{y_i} \|^2_2,
\end{equation}
This means that minimizing $\Tilde{\mathcal{L}}$ is equivalent to minimizing a non-quantized neural network while regularizing the norm of the gradients.
\end{theorem*}
From Eq. \ref{eq:mse_quant}, we have,
\begin{equation*}
\Tilde{\mathcal{L}} = \frac{1}{m} \sum_{i=1}^m \left[ \|{y_i}\|_2^2 + \|\hat{y_i}^q\|_2^2 - 2 {y_i} \hat{y_i}^q
\right] = \mathbb{E}_{p(x, y, \delta)} \left[ \|{y_i}\|_2^2 + \|\hat{y_i}^q\|_2^2 - 2 {y_i} \hat{y_i}^q
\right]
\end{equation*}

Also, since $\hat{y_i}^q = f(x, w+\delta)$, we can apply a first order Taylor approximation,
\begin{equation*}
f(x, w+\delta) = f(x, w) + \delta \nabla_w f(x, w) + \mathcal{O}(\delta^2)
\end{equation*}

Thus, $\hat{y_i}^q \approx \hat{y_i} + \delta \nabla_w \hat{y_i}$.

Re-writing, the expectation on $\Tilde{\mathcal{L}}$, 
\begin{align*}
    \Tilde{\mathcal{L}} &= \mathbb{E}_{p(x, y, \delta)} \left[ \|{y_i}\|_2^2 + \|\hat{y_i}^q\|_2^2 - 2 {y_i} \hat{y_i}^q \right]\\
    &\approx \mathbb{E}_{p(x, y, \delta)} \left[ \|{y_i}\|_2^2 + \| \hat{y_i} + \delta \nabla_w \hat{y_i}\|_2^2  - 2 y_i (\hat{y_i} + \delta \nabla_w \hat{y_i}) \right]\\
    &= \mathbb{E}_{p(x, y, \delta)} \left[ \|{y_i}\|_2^2 + \|\hat{y_i}\|_2^2 - 2 {y_i} \hat{y_i} + \delta^2 \|\nabla_w \hat{y_i} \|^2_2 + 2\delta (\nabla_w \hat{y_i}) (\hat{y} - y) \right]\\
    &= \mathcal{L} + \sigma \delta^2 \mathbb{E}_{p(x, y, \delta)}  [\|\nabla_w \hat{y_i} \|^2_2]
\end{align*}

Note that since $\delta \sim \mathcal{N} (0, \sigma I)$, we have $\mathbb{E}_{p(x, y, \delta)} [2\delta (\nabla_w \hat{y_i}) (\hat{y} - y)] = 0$.

\newpage
\section{Extended Results}
\label{appendix:b}

\begin{figure}[!h]
\centering
    \subfigure[]{
            \includegraphics[width=6cm]{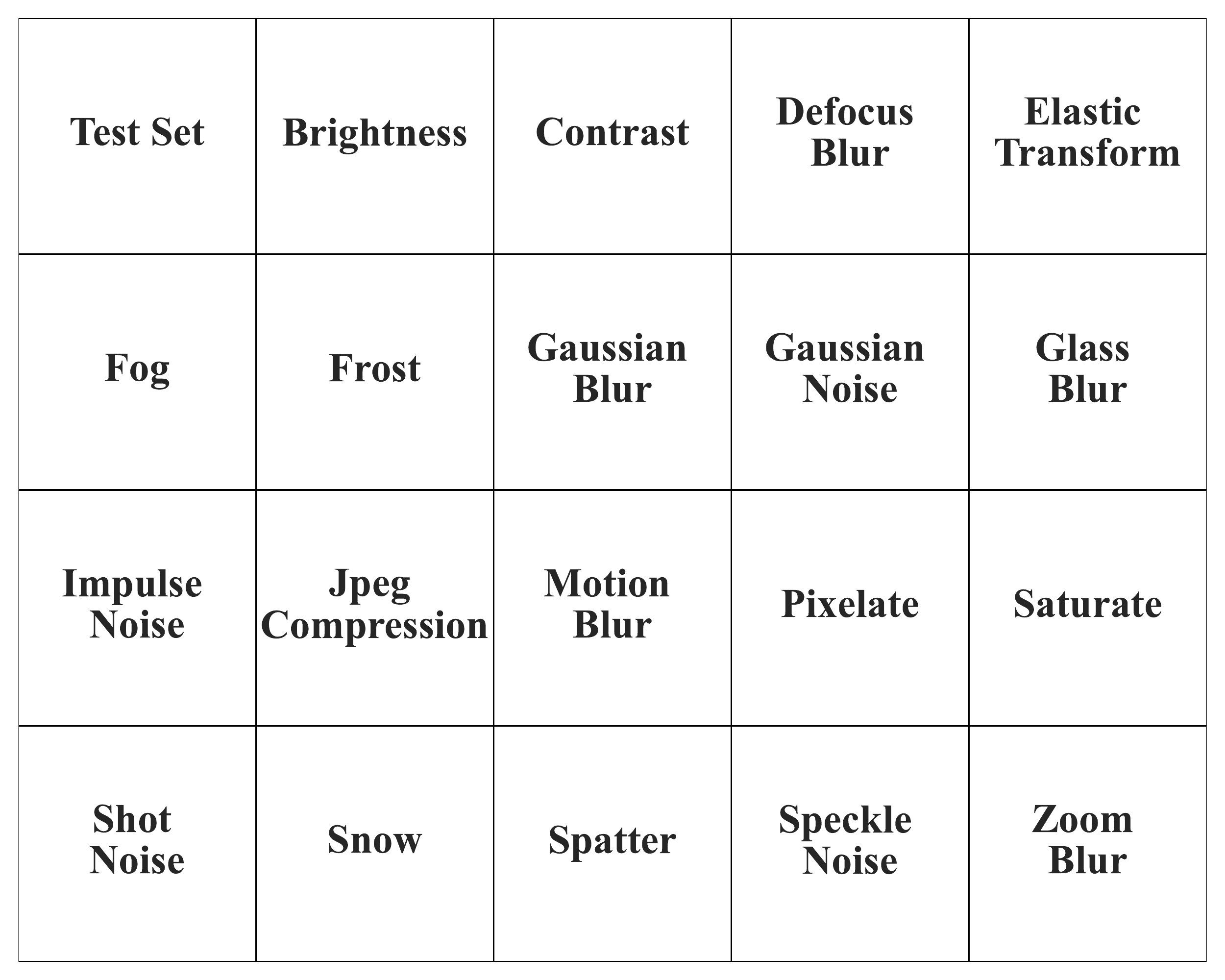}
            \label{fig:aug_map}
    }
    \subfigure[]{
            \includegraphics[width=6cm]{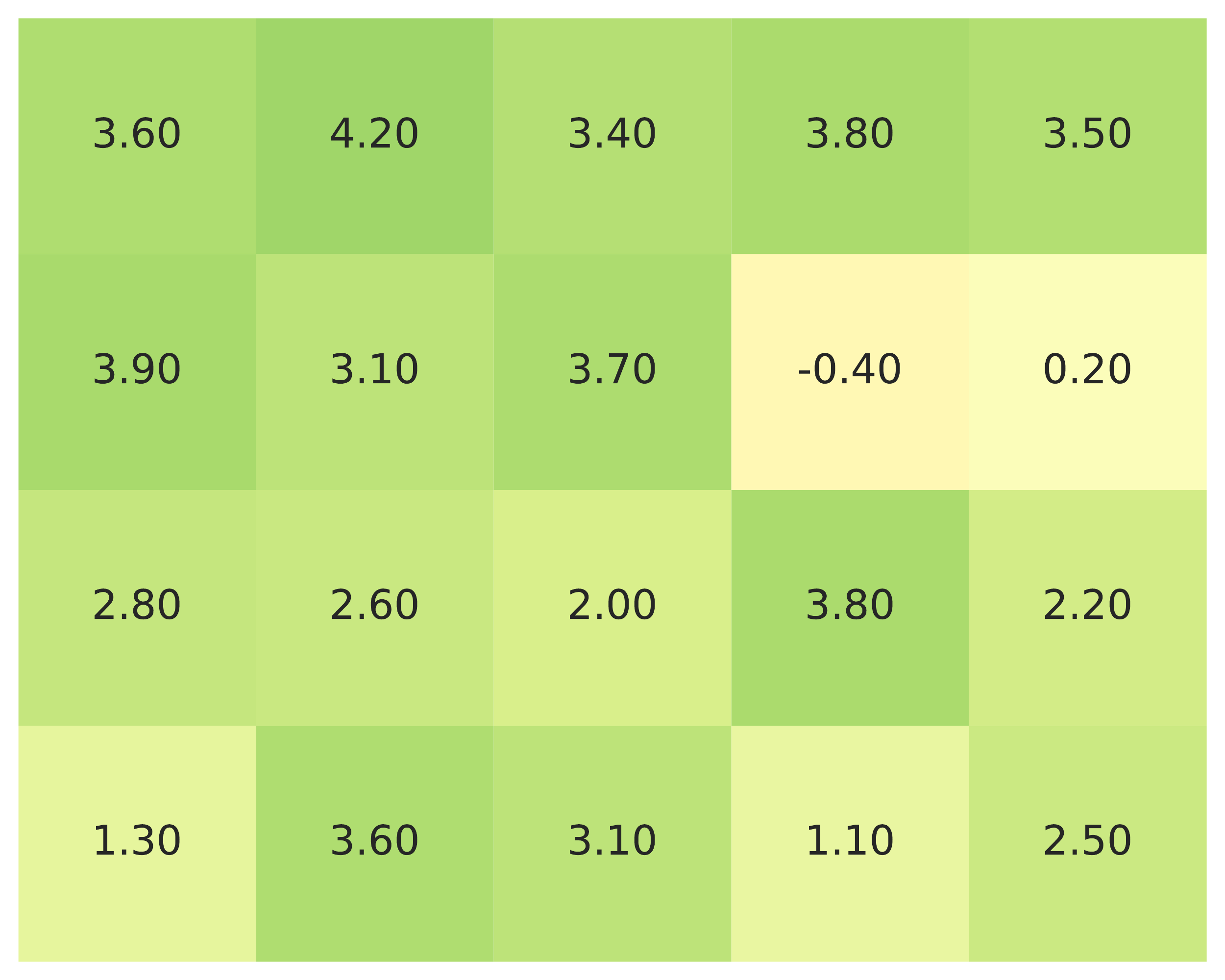}
            \label{fig:res_18_exa}
    }
\caption{ Figure \ref{fig:aug_map} illustrates the positional location of augmentation that was applied on the original test dataset. In addition to test images for each dataset, in our experiments, we applied 19 different augmentations on the test dataset. We used the augmentation that is provided in \cite{hendrycks2019robustness}. Figure \ref{fig:res_18_exa} illustrates the generalization improvement between quantized and full precision Resnet18 on CIFAR100 model when different augmentation is applied. The value in each cell shows the difference and the position of each cell corresponds to the data augmentation technique listed in  Figure \ref{fig:aug_map}. }
\end{figure}

\vspace{5cm}

\begin{figure}[!h]
\label{fig:all_result}
  \begin{picture}(0,200)
    \put(30,300){2 bits}
    \put(110,300){4 bits}
    \put(190,300){8 bits}

    \put(280,300){2 bits}
    \put(350,300){4 bits}
    \put(430,300){8 bits}
    \put(-10,240){\rotatebox{90}{Resnet20}}
    \put(-10,150){\rotatebox{90}{Mobilenet V1}}
    \put(-10,90){\rotatebox{90}{Resnet18}}
    \put(-10,20){\rotatebox{90}{Resnet50}}

    \put(3, -12){\rule[1pt]{1pt}{4pt}}
    \put(3, -12){\rule[1pt]{238pt}{1pt}}
    \put(80, -8){LSQ Quantization}
    \put(240, -12){\rule[1pt]{1pt}{4pt}}

    \put(250, -12){\rule[1pt]{1pt}{4pt}}
    \put(250, -12){\rule[1pt]{238pt}{1pt}}
    \put(300, -8){PACT + DoReFa Quantization}
    \put(487, -12){\rule[1pt]{1pt}{4pt}}
    
    \includegraphics[width=\textwidth/2]{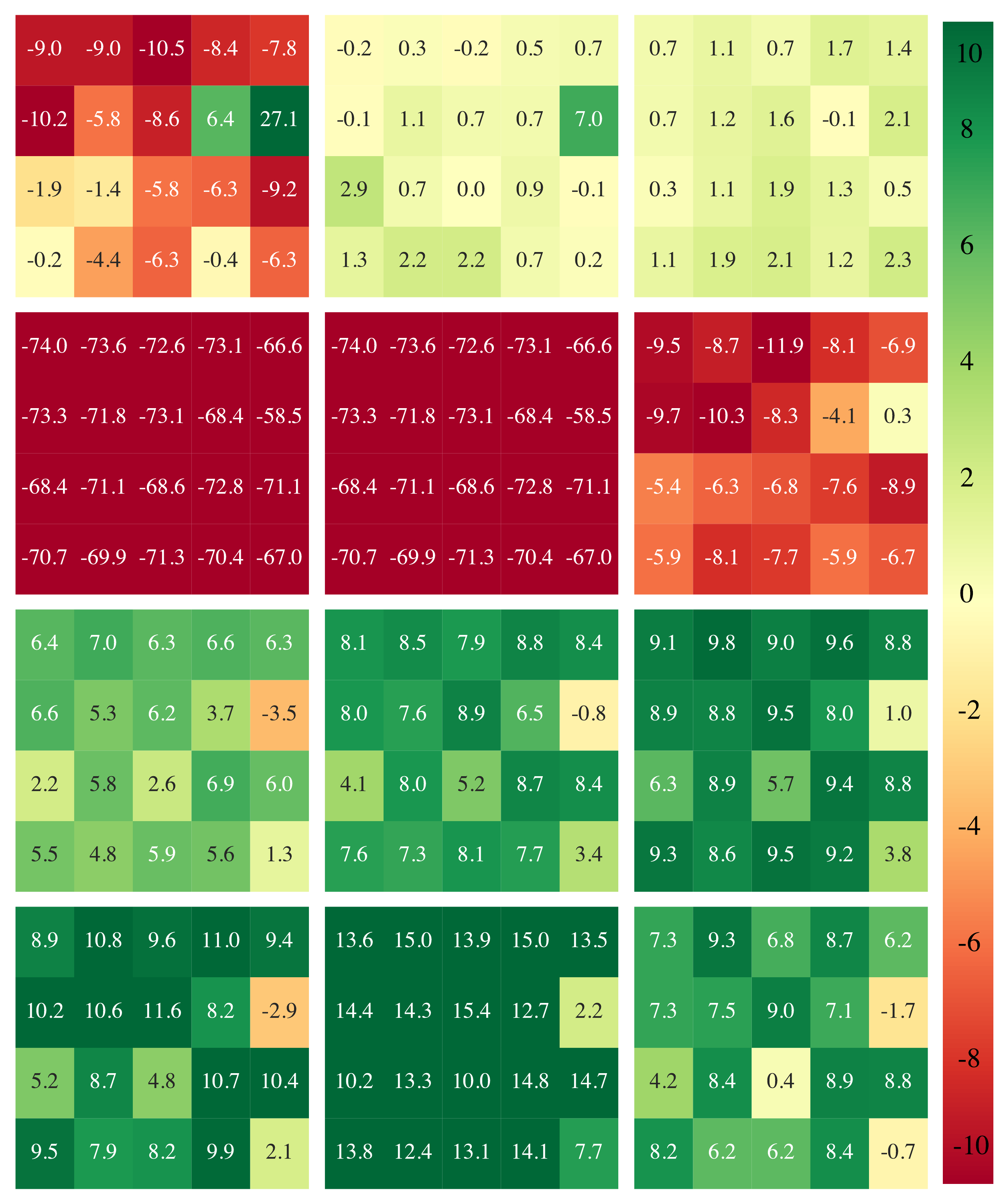}
    \hspace{1mm}
    \includegraphics[width=\textwidth/2]{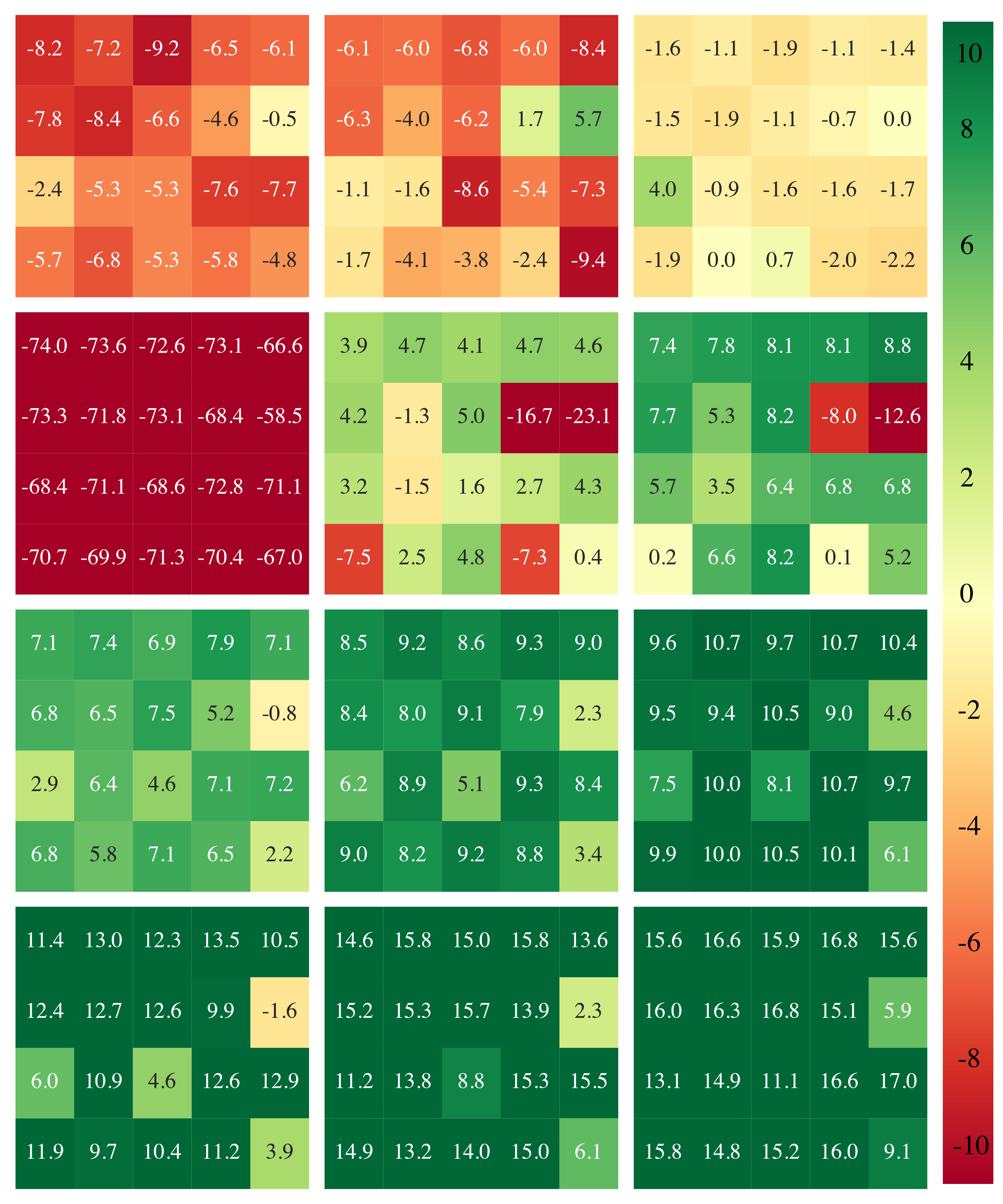}

  \end{picture}

  \caption{Regularization effect of quantization in classification task when different quantization methods and levels are used on CIFAR10 dataset.}
\end{figure}

\newpage

\begin{figure}[!h]
\label{fig:all_result}
\vspace{1cm}
  \begin{picture}(0,200)
    \put(30,220){2 bits}
    \put(110,220){4 bits}
    \put(190,220){8 bits}

    \put(280,220){2 bits}
    \put(350,220){4 bits}
    \put(430,220){8 bits}
    \put(-10,150){\rotatebox{90}{Mobilenet V1}}
    \put(-10,90){\rotatebox{90}{Resnet18}}
    \put(-10,20){\rotatebox{90}{Resnet50}}

    \put(3, -12){\rule[1pt]{1pt}{4pt}}
    \put(3, -12){\rule[1pt]{238pt}{1pt}}
    \put(80, -8){LSQ Quantization}
    \put(240, -12){\rule[1pt]{1pt}{4pt}}

    \put(250, -12){\rule[1pt]{1pt}{4pt}}
    \put(250, -12){\rule[1pt]{238pt}{1pt}}
    \put(300, -8){PACT + DoReFa Quantization}
    \put(487, -12){\rule[1pt]{1pt}{4pt}}
    
    \includegraphics[width=\textwidth/2]{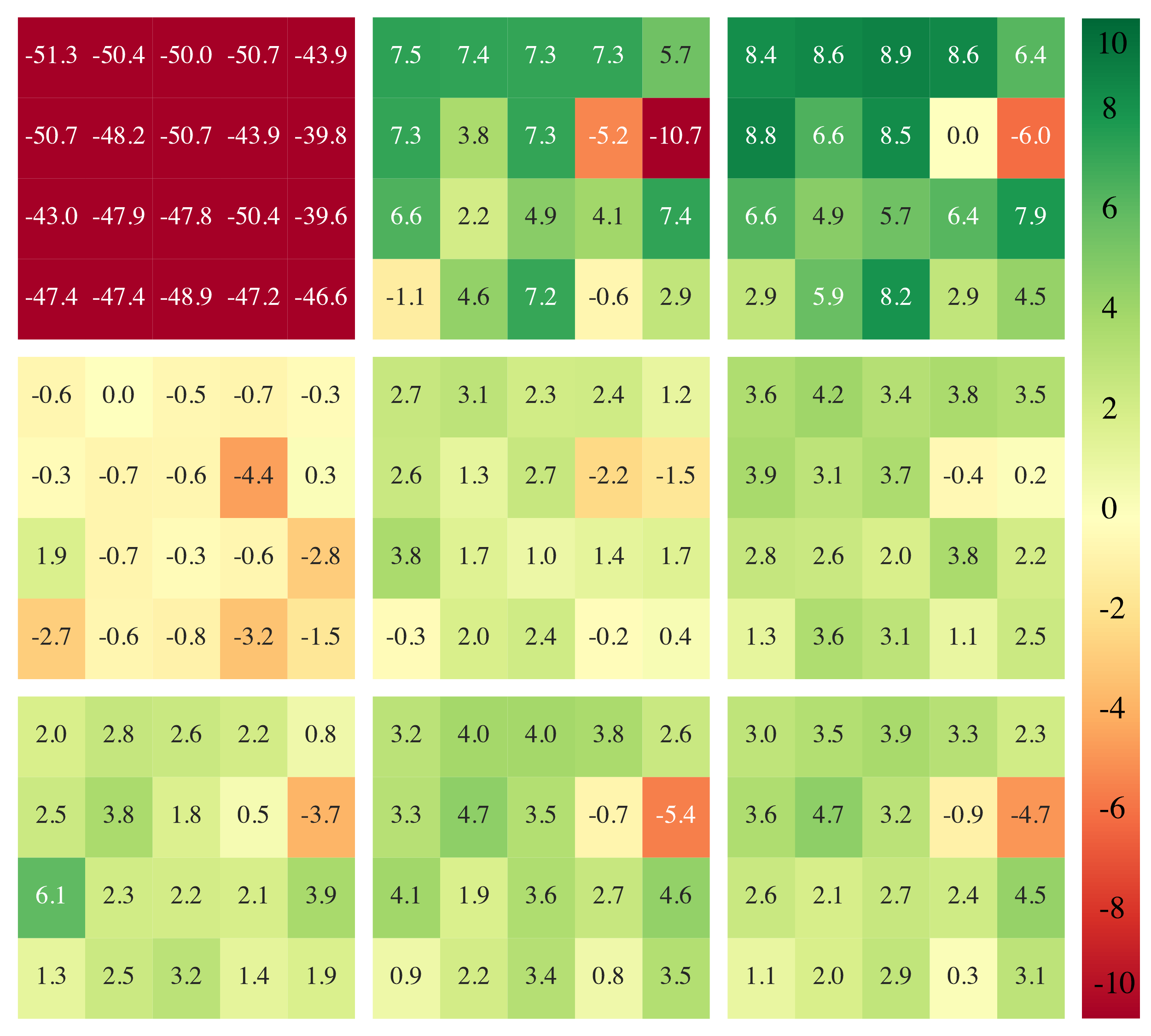}
    \hspace{1mm}
    \includegraphics[width=\textwidth/2]{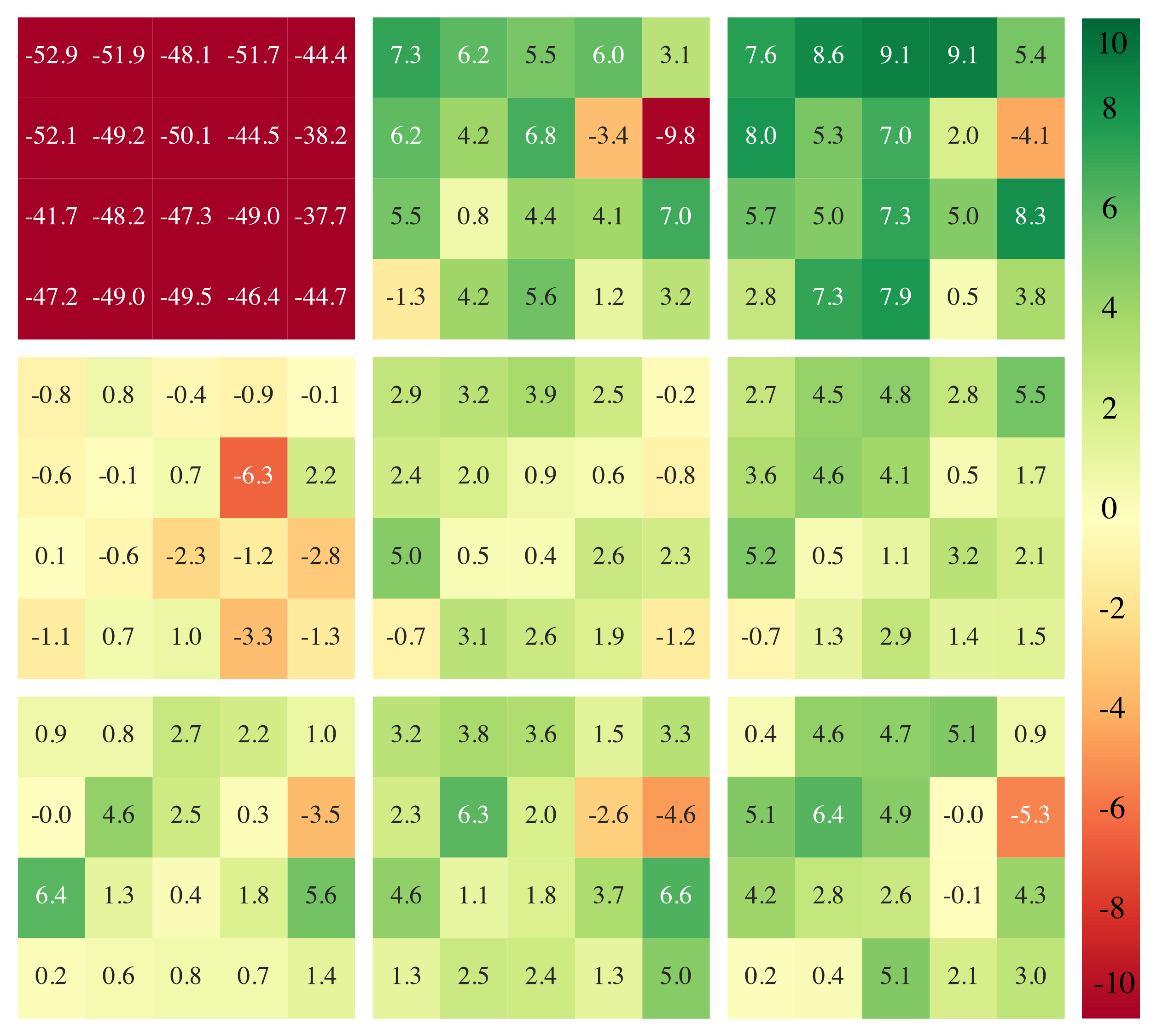}

  \end{picture}

  \caption{Regularization effect of quantization in classification task when different quantization methods and levels are used on CIFAR100 dataset.}
\end{figure}

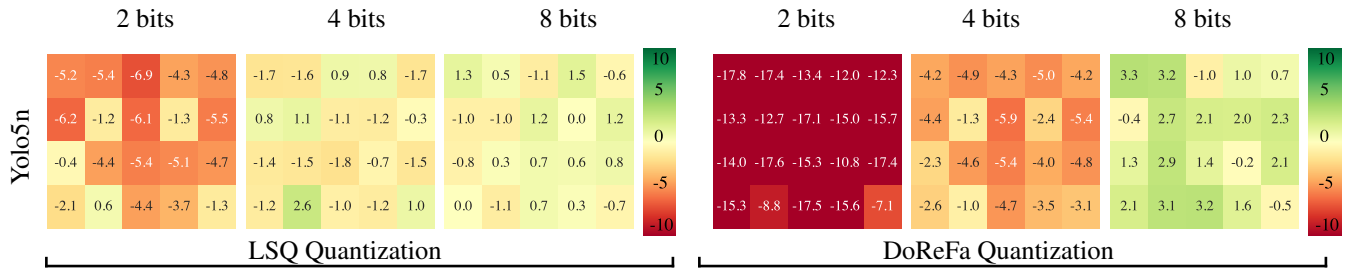
\begin{figure}[!h]
\label{fig:all_result}
\vspace{-4cm}
  \begin{picture}(0,200)
    \put(30,80){2 bits}
    \put(110,80){4 bits}
    \put(190,80){8 bits}

    \put(280,80){2 bits}
    \put(350,80){4 bits}
    \put(430,80){8 bits}
    \put(-10,20){\rotatebox{90}{Yolo5n}}

    \put(3, -12){\rule[1pt]{1pt}{4pt}}
    \put(3, -12){\rule[1pt]{238pt}{1pt}}
    \put(80, -8){LSQ Quantization}
    \put(240, -12){\rule[1pt]{1pt}{4pt}}

    \put(250, -12){\rule[1pt]{1pt}{4pt}}
    \put(250, -12){\rule[1pt]{238pt}{1pt}}
    \put(320, -8){DoReFa Quantization}
    \put(487, -12){\rule[1pt]{1pt}{4pt}}
    
    \includegraphics[width=\textwidth/2]{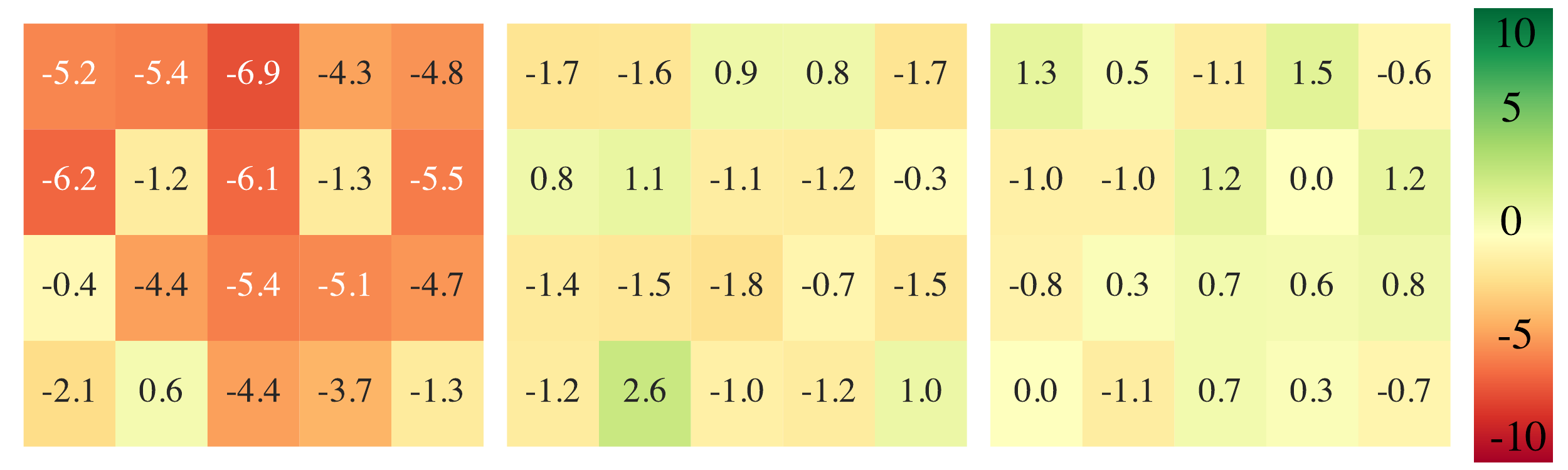}
    \hspace{1mm}
    \includegraphics[width=\textwidth/2]{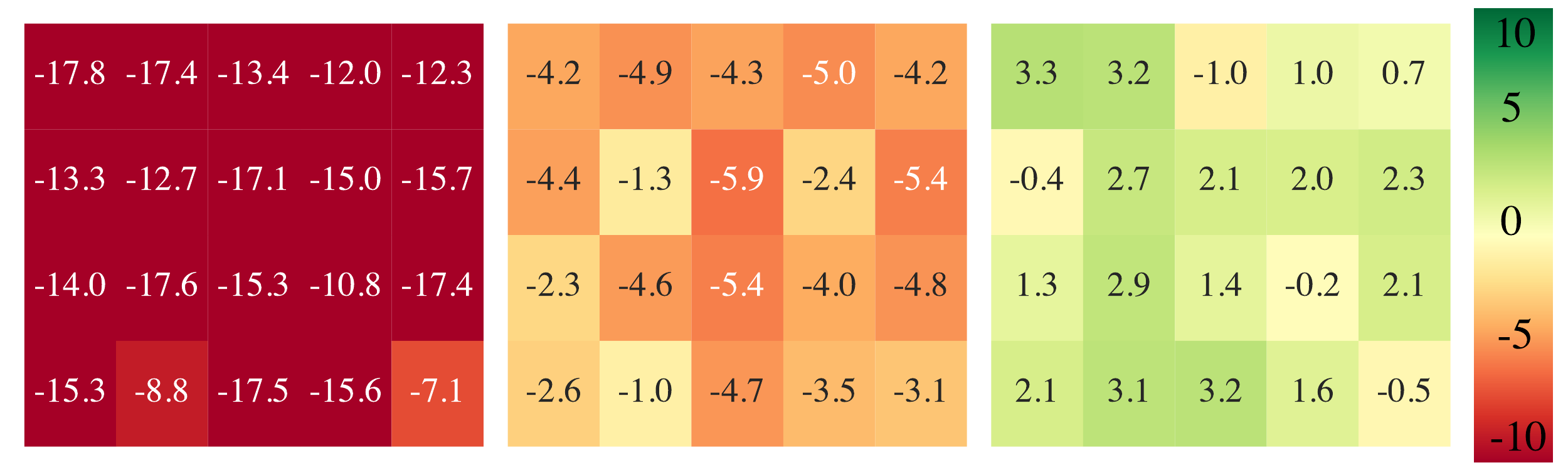}

  \end{picture}

  \caption{Regularization effect of quantization in object detection task when different quantization methods and levels are used. NOTE: Unlike previous results we only quantized weights (we used DoReFa and not PACT+DoReFa) for right column results.}
\end{figure}



  

\end{document}